\documentclass{article}

    \PassOptionsToPackage{numbers, compress}{natbib}

    \usepackage[preprint]{neurips_2021}

\usepackage[utf8]{inputenc} 
\usepackage[T1]{fontenc}    
\usepackage{hyperref}       
\usepackage{url}            
\usepackage{booktabs}       
\usepackage{amsfonts}       
\usepackage{nicefrac}       
\usepackage{microtype}      
\usepackage{xcolor}         

\usepackage{makecell}
\usepackage{multirow}
\usepackage{epsfig}
\usepackage{amsmath}
\usepackage{amssymb}

\usepackage{wrapfig}

\newcommand{\demph}[1]{\textcolor[rgb]{0.565,0.565,0.565}{#1}}

\title{Video Swin Transformer}

\author{%
  Ze Liu\thanks{~Equal Contribution. $^{\dag}$ Equal Advising. The work is done when Ze Liu, Jia Ning and Yixuan Wei are interns at Microsoft Research Asia. }\hspace{1.2mm}$^{12}$, Jia Ning$^{*13}$, Yue Cao$^{1\dag}$, Yixuan Wei$^{14}$, Zheng Zhang$^{1}$, Stephen Lin$^{1}$, Han Hu$^{1\dag}$ \\
  $^1$Microsoft Research Asia\\
  $^2$University of Science and Technology of China\\
  $^3$Huazhong University of Science and Technology\\
  $^4$Tsinghua University\\
}

\begin{document}

\maketitle

\begin{abstract}
The vision community is witnessing a modeling shift from CNNs to Transformers, where pure Transformer architectures have attained top accuracy on the major video recognition benchmarks. These video models are all built on Transformer layers that globally connect patches across the spatial and temporal dimensions. In this paper, we instead advocate an inductive bias of locality in video Transformers, which leads to a better speed-accuracy trade-off compared to previous approaches which compute self-attention globally even with spatial-temporal factorization. The locality of the proposed video architecture is realized by adapting the Swin Transformer designed for the image domain, while continuing to leverage the power of pre-trained image models. Our approach achieves state-of-the-art accuracy on a broad range of video recognition benchmarks, including on action recognition (84.9 top-1 accuracy on Kinetics-400 and 86.1 top-1 accuracy on Kinetics-600 with $\sim$20$\times$ less pre-training data and $\sim$3$\times$ smaller model size) and temporal modeling (69.6 top-1 accuracy on Something-Something v2). The code and models will be made publicly available at~\url{https://github.com/SwinTransformer/Video-Swin-Transformer}.
\end{abstract}

\section{Introduction}

Convolution-based backbone architectures have long dominated visual modeling in computer vision~\cite{lecun1998lenet,krizhevsky2012alexnet,simonyan2014vgg,szegedy2015googlenet,he2015resnet,huang2017densely}. However, a modeling shift is currently underway on backbone architectures for image classification, from Convolutional Neural Networks (CNNs) to Transformers~\cite{dosovitskiy2020vit,touvron2020deit,liu2021swin}. This trend began with the introduction of Vision Transformer (ViT)~\cite{dosovitskiy2020vit,touvron2020deit}, which globally models spatial relationships on non-overlapping image patches with the standard Transformer encoder~\cite{vaswani2017attention}. The great success of ViT on images has led to investigation of Transformer-based architectures for video-based recognition tasks~\cite{arnab2021vivit, timesformer2021}.

Previously for convolutional models, backbone architectures for video were adapted from those for images simply by extending the modeling through the temporal axis. For example, 3D convolution~\cite{tran2015learning} is a direct extension of 2D convolution for joint spatial and temporal modeling at the operator level. As joint spatiotemporal modeling is not economical or easy to optimize, factorization of the spatial and temporal domains was proposed to achieve a better speed-accuracy tradeoff~\cite{qiu2017P3D,xie2018rethinking}. In the initial attempts at Transformer-based video recognition, a factorization approach is also employed, via a factorized encoder~\cite{arnab2021vivit} or factorized self-attention~\cite{arnab2021vivit, timesformer2021}. This has been shown to greatly reduce model size without a substantial drop in performance.

In this paper, we present a pure-transformer backbone architecture for video recognition that is found to surpass the factorized models in efficiency. It achieves this by taking advantage of the inherent spatiotemporal locality of videos, in which pixels that are closer to each other in spatiotemporal distance are more likely to be correlated. Because of this property, full spatiotemporal self-attention can be well-approximated by self-attention computed locally, at a significant saving in computation and model size. 

We implement this approach through a spatiotemporal adaptation of Swin Transformer~\cite{liu2021swin}, which was recently introduced as a general-purpose vision backbone for image understanding. Swin Transformer incorporates inductive bias for spatial locality, as well as for hierarchy and translation invariance. Our model, called Video Swin Transformer, strictly follows the hierarchical structure of the original Swin Transformer, but extends the scope of local attention computation from only the spatial domain to the spatiotemporal domain. As the local attention is computed on non-overlapping windows, the shifted window mechanism of the original Swin Transformer is also reformulated to process spatiotemporal input.

As our architecture is adapted from Swin Transformer, it can readily be initialized with a strong model pre-trained on a large-scale image dataset.
With a model pre-trained on ImageNet-21K, we interestingly find that the learning rate of the backbone architecture needs to be smaller (e.g. 0.1$\times$) than that of the head, which is randomly initialized. As a result, the backbone forgets the pre-trained parameters and data slowly while fitting the new video input, leading to better generalization. This observation suggests a direction for further study on how to better utilize pre-trained weights.

The proposed approach shows strong performance on the video recognition tasks of action recognition on Kinetics-400/Kinetics-600 and temporal modeling on Something-Something v2 (abbreviated as SSv2).
For video action recognition, its 84.9\% top-1 accuracy on Kinetics-400 and 86.1\% top-1 accuracy on Kinetics-600 slightly surpasses the previous state-of-the-art results (ViViT~\cite{arnab2021vivit}) by +0.1/+0.3 points, with a smaller model size (200.0M params for Swin-L vs.~647.5M params for ViViT-H) and a smaller pre-training dataset (ImageNet-21K vs.~JFT-300M).
For temporal modeling on SSv2, it obtains 69.6\% top-1 accuracy, an improvement of +0.9 points over previous state-of-the-art (MViT~\cite{mvit2021}).

\section{Related Works}

\paragraph{CNN and variants} 
In computer vision, convolutional networks have long been the standard for backbone architectures.
For 3D modeling, C3D~\cite{tran2015learning} is a pioneering work that devises a 11-layer deep network with 3D convolutions. 
The work on I3D~\cite{carreira2017i3d} reveals that inflating the 2D convolutions in Inception V1 to 3D convolutions, with initialization by ImageNet pretrained weights, achieves good results on large-scale Kinetics datasets.
In P3D~\cite{qiu2017P3D}, S3D~\cite{xie2018rethinking} and R(2+1)D~\cite{tran2018closer}, it is found that disentangling spatial and temporal convolution leads to a speed-accuracy tradeoff better than the original 3D convolution.
The potential of convolution based approaches is limited by the small receptive field of the convolution operator. With a self-attention mechanism, the receptive field can be broadened with fewer parameters and lower computation costs, which leads to better performance of vision Transformers on video recognition.

\paragraph{Self-attention/Transformers to complement CNNs}
NLNet~\cite{wang2018non} is the first work to adopt self-attention to model pixel-level long-range dependency for visual recognition tasks. 
GCNet~\cite{cao2019gcnet} presents an observation that the accuracy improvement of NLNet can mainly be ascribed to its global context modeling, and thus it simplifies the NL block into a lightweight global context block which matches NLNet in performance but with fewer parameters and less computation.
DNL~\cite{yin2020DNL} on the contrary attempts to alleviate this degeneration problem by a disentangled design that allows learning of different contexts for different pixels while preserving the shared global context.
All these approaches provide a complementary component to CNNs for modeling long range dependency. In our work, we show that a pure-transformer based approach more fully captures the power of self-attention, leading to superior performance.

\paragraph{Vision Transformers}
A shift in backbone architectures for computer vision, from CNNs to Transformers, began recently with Vision Transformer (ViT)~\cite{dosovitskiy2020vit,touvron2020deit}. This seminal work has led to subsequent research that aims to improve its utility. 
DeiT~\cite{touvron2020deit} integrates several training strategies that allow ViT to also be effective using the smaller ImageNet-1K dataset. 
Swin Transformer~\cite{liu2021swin} further introduces the inductive biases of locality, hierarchy and translation invariance, which enable it to serve as a general-purpose backbone for various image recognition tasks.

The great success of image Transformers has led to investigation of Transformer-based architectures for video-based recognition tasks~\cite{neimark2021VTN,arnab2021vivit, timesformer2021,mvit2021,li2021vidtr}.
VTN~\cite{neimark2021VTN} proposes to add a temporal attention encoder on top of the pre-trained ViT, which yields good performance on video action recognition.
TimeSformer~\cite{timesformer2021} studies five different variants of space-time attention and suggests a factorized space-time attention for its strong speed-accuracy tradeoff.
ViViT~\cite{arnab2021vivit} examines four factorized designs of spatial and temporal attention for the pre-trained ViT model, and suggests an architecture similar to VTN that achieves state-of-the-art performance on the Kinetics dataset.
MViT~\cite{mvit2021} is a multi-scale vision transformer for video recognition trained from scratch that reduces computation by pooling attention for spatiotemporal modeling, which leads to state-of-the-art results on SSv2.
All these studies are based on global self-attention modules. In this paper, we first investigate spatiotemporal locality and then empirically show that the Video Swin Transformer with spatiotemporal locality bias surpasses the performance of all the other vision Transformers on various video recognition tasks.

\section{Video Swin Transformer}

\begin{figure}
    \centering
    \includegraphics[width=1.\linewidth]{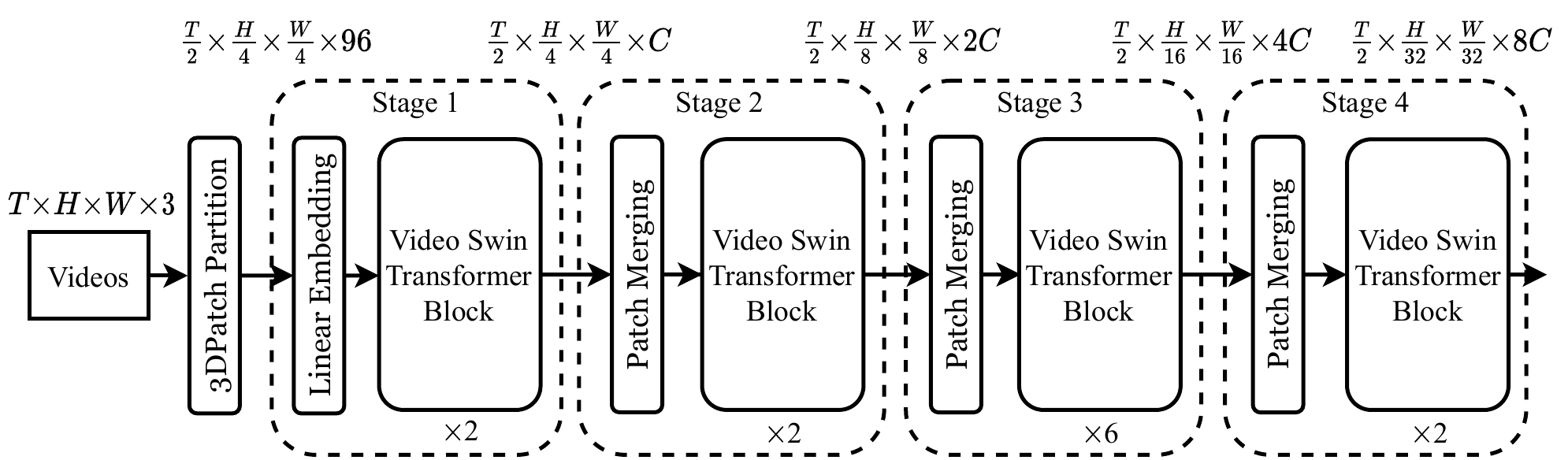}
    \caption{Overall architecture of Video Swin Transformer (tiny version, referred to as Swin-T).}
    \label{fig:arch}
\end{figure}

\subsection{Overall Architecture}

The overall architecture of the proposed Video Swin Transformer is shown in Figure~\ref{fig:arch}, which illustrates its tiny version (Swin-T). 
The input video is defined to be of size $T$$\times$$H$$\times$$W$$\times$$3$, consisting of $T$ frames which each contain $H$$\times$$W$$\times$$3$ pixels.
In Video Swin Transformer, we treat each 3D patch of size 2$\times$4$\times$4$\times$3 as a token.
Thus, the 3D patch partitioning layer obtains $\frac{T}{2}$$\times$$\frac{H}{4}$$\times$$\frac{W}{4}$ 3D tokens, with each patch/token consisting of a 96-dimensional feature. 
A linear embedding layer is then applied to project the features of each token to an arbitrary dimension denoted by $C$.

\begin{wrapfigure}{r}{0.38\linewidth}
    \centering
    \includegraphics[width=1.0\linewidth]{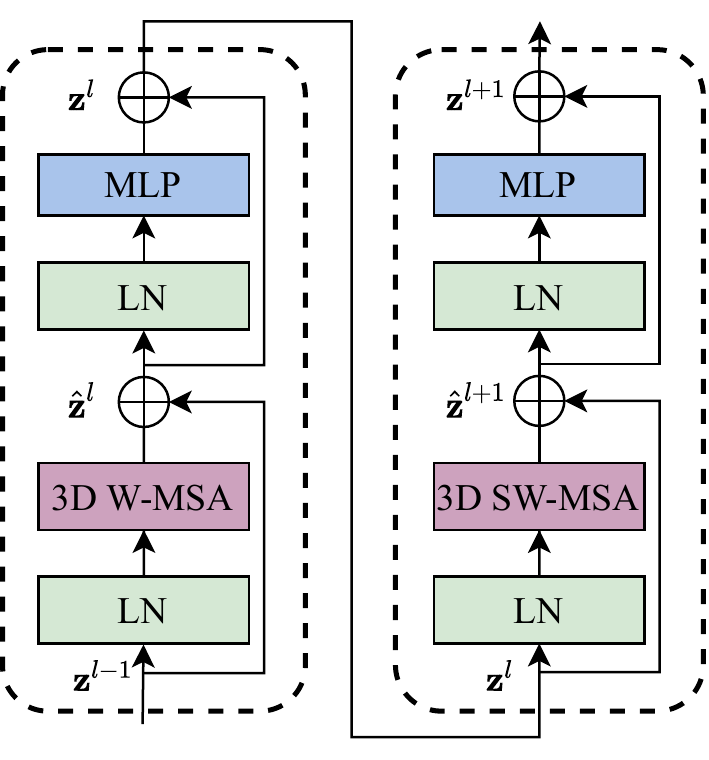}
    \caption{An illustration of two successive Video Swin Transformer blocks.}
    \label{fig:blk}
\end{wrapfigure}

Following the prior art~\cite{qiu2017P3D,xie2018rethinking,feichtenhofer2019slowfast,feichtenhofer2020x3d}, we do not downsample along the temporal dimension. This allows us to strictly follow the hierarchical architecture of the original Swin Transformer~\cite{liu2021swin}, which consists of four stages and performs 2$\times$ spatial downsampling in the patch merging layer of each stage. The patch merging layer concatenates the features of each group of 2$\times$2 spatially neighboring patches and applies a linear layer to project the concatenated features to half of their dimension. For example, the linear layer in the second stage projects $4C$-dimensional features for each token to $2C$ dimensions.

The major component of the architecture is the Video Swin Transformer block, which is built by replacing the multi-head self-attention (MSA) module in the standard Transformer layer with the 3D shifted window based multi-head self-attention module (presented in Section~\ref{sec:att}) and keeping the other components unchanged.
Specifically, a video transformer block consists of a 3D shifted window based MSA module followed by a feed-forward network, specifically a 2-layer MLP, with GELU non-linearity in between. Layer Normalization (LN) is applied before each MSA module and FFN, and a residual connection is applied after each module.
The computational formulas of the Video Swin Transformer block are given in Eqn.~\eqref{eq.swin}.

\subsection{3D Shifted Window based MSA Module}
Compared to images, videos require a much larger number of input tokens to represent them, as videos additionally have a temporal dimension. A global self-attention module would thus be unsuitable for video tasks as this would lead to enormous computation and memory costs. Here, we follow Swin Transformer by introducing a locality inductive bias to the self-attention module, which is later shown to be effective for video recognition.

\begin{figure}
    \centering
    \includegraphics[width=1.\linewidth]{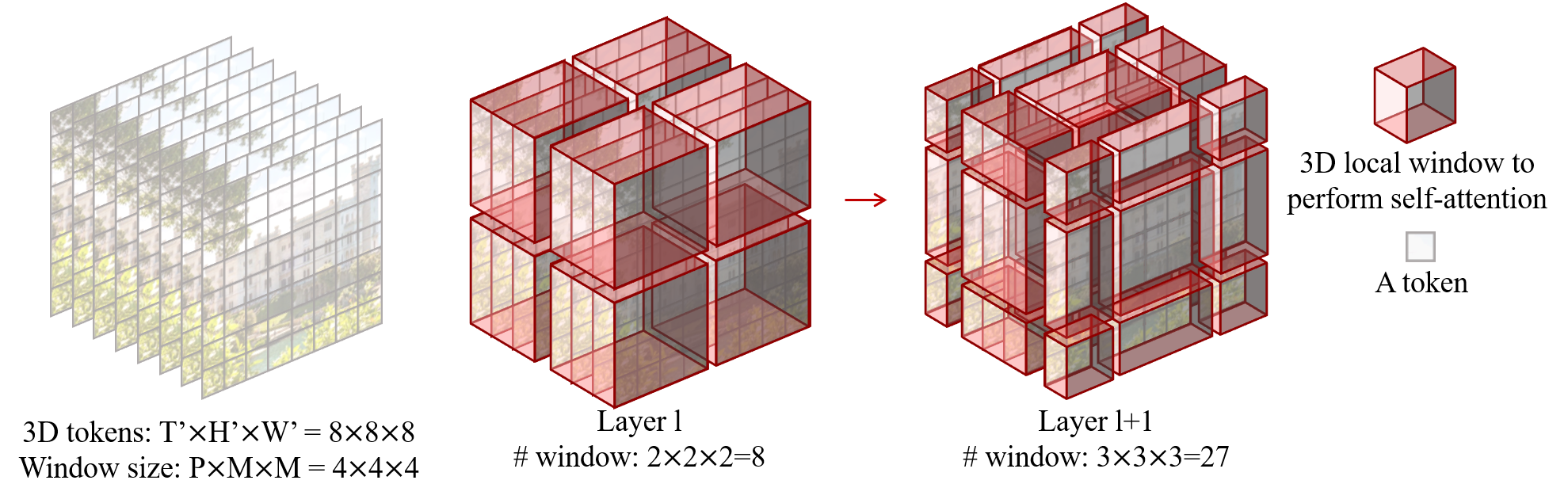}
    \caption{An illustrated example of 3D shifted windows. The input size $T'$$\times$$H'$$\times$$W'$ is 8$\times$8$\times$8, and the 3D window size $P$$\times$$M$$\times$$M$ is 4$\times$4$\times$4. As layer $l$ adopts regular window partitioning, the number of windows in layer $l$ is 2$\times$2$\times$2=8. For layer $l$+$1$, as the windows are shifted by ($\frac{P}{2}$,$\frac{M}{2}$,$\frac{M}{2}$)=(2, 2, 2) tokens, the number of windows becomes 3$\times$3$\times$3=27. Though the number of windows is increased, the efficient batch computation in~\cite{liu2021swin} for the shifted configuration can be followed, such that the final number of windows for computation is still 8.}
    \label{fig:shift-window}
\end{figure}

\paragraph{Multi-head self-attention on non-overlapping 3D windows} 
Multi-head self-attention (MSA) mechanisms on each non-overlapping 2D window has been shown to be both effective and efficient for image recognition. Here, we straightforwardly extend this design to process video input. Given a video composed of $T'$$\times$$H'$$\times$$W'$ 3D tokens and a 3D window size of $P$$\times$$M$$\times$$M$, the windows are arranged to evenly partition the video input in a non-overlapping manner. That is, the input tokens are partitioned into $\lceil\frac{T'}{P}\rceil$$\times$$\lceil\frac{H'}{M}\rceil$$\times$$\lceil\frac{W'}{M}\rceil$ non-overlapping 3D windows.
For example, as shown in Figure~\ref{fig:shift-window}, for an input size of 8$\times$8$\times$8 tokens and a window size of 4$\times$4$\times$4, the number of windows in layer $l$ would be 2$\times$2$\times$2=8.
And the multi-head self-attention is performed within each 3D window.

\paragraph{3D Shifted Windows}\label{sec:att}
As the multi-head self-attention mechanism is applied within each non-overlapping 3D window, there lacks connections across different windows, which may limit the representation power of the architecture. Thus, we extend the shifted 2D window mechanism of Swin Transformer to 3D windows for the purpose of introducing cross-window connections while maintaining the efficient computation of non-overlapping window based self-attention.

Given that the number of input 3D tokens is $T'$$\times$$H'$$\times$$W'$ and the size of each 3D window is $P$$\times$$M$$\times$$M$, for two consecutive layers, the self-attention module in the first layer uses the regular window partition strategy such that we obtain $\lceil\frac{T'}{P}\rceil$$\times$$\lceil\frac{H'}{M}\rceil$$\times$$\lceil\frac{W'}{M}\rceil$ non-overlapping 3D windows. For the self-attention module in the second layer, the window partition configuration is shifted along the temporal, height and width axes by ($\frac{P}{2}$,$\frac{M}{2}$,$\frac{M}{2}$) tokens from that of the preceding layer's self-attention module.

We illustrate this with an example in Figure~\ref{fig:shift-window}. The input size is 8$\times$8$\times$8, and the window size is 4$\times$4$\times$4. As layer $l$ adopts regular window partitioning, the number of windows in layer $l$ is 2$\times$2$\times$2=8. For layer $l+1$, as the windows are shifted by ($\frac{P}{2}$,$\frac{M}{2}$,$\frac{M}{2}$)=(2, 2, 2) tokens, the number of windows becomes 3$\times$3$\times$3=27. Though the number of windows is increased, the efficient batch computation in~\cite{liu2021swin} for the shifted configuration can be followed, such that the final number of windows for computation is still 8.

With the shifted window partitioning approach, two consecutive Video Swin Transformer blocks are computed as
\begin{align}
    &{{\hat{\bf{z}}}^{l}} = \text{3DW-MSA}\left( {\text{LN}\left( {{{\bf{z}}^{l - 1}}} \right)} \right) + {\bf{z}}^{l - 1},\nonumber\\
    &{{\bf{z}}^l} = \text{FFN}\left( {\text{LN}\left( {{{\hat{\bf{z}}}^{l}}} \right)} \right) + {{\hat{\bf{z}}}^{l}},\nonumber\\
    &{{\hat{\bf{z}}}^{l+1}} = \text{3DSW-MSA}\left( {\text{LN}\left( {{{\bf{z}}^{l}}} \right)} \right) + {\bf{z}}^{l}, \nonumber\\
    &{{\bf{z}}^{l+1}} = \text{FFN}\left( {\text{LN}\left( {{{\hat{\bf{z}}}^{l+1}}} \right)} \right) + {{\hat{\bf{z}}}^{l+1}}, \label{eq.swin}
\end{align}
where ${\hat{\bf{z}}}^l$ and ${\bf{z}}^l$ denote the output features of the 3D(S)W-MSA module and the FFN module for block $l$, respectively; $\text{3DW-MSA}$ and $\text{3DSW-MSA}$ denote 3D window based multi-head self-attention using regular and shifted window partitioning configurations, respectively.

Similar to image recognition~\cite{liu2021swin}, this 3D shifted window design introduces connections between neighboring non-overlapping 3D windows in the previous layer. This will later be shown to be effective for several video recognition tasks, such as action recognition on Kinetics 400/600 and temporal modeling on SSv2.

\paragraph{3D Relative Position Bias} Numerous previous works~\cite{raffel2019t5,bao2020unilmv2,hu2018relation,hu2019localrelation} have shown that it can be advantageous to include a relative position bias to each head in self-attention computation. Thus, we follow~\cite{liu2021swin} by introducing 3D relative position bias $B \in \mathbb{R}^{P^2 \times M^2 \times M^2}$ for each head as
\begin{equation}
\label{eq.att}
    \text{Attention}(Q, K, V) = \text{SoftMax}(QK^T/\sqrt{d}+B)V,
\end{equation}
where $Q, K, V \in \mathbb{R}^{PM^2\times d}$ are the \emph{query}, \emph{key} and \emph{value} matrices; $d$ is the dimension of \emph{query} and \emph{key} features, and $PM^2$ is the number of tokens in a 3D window. Since the relative position along each axis lies in the range of  $[-P+1, P-1]$ (temporal) or $[-M+1, M-1]$ (height or width), we parameterize a smaller-sized bias matrix $\hat{B} \in \mathbb{R}^{(2P-1)\times(2M-1)\times (2M-1)}$, and values in $B$ are taken from $\hat{B}$.

\subsection{Architecture Variants}

Following~\cite{liu2021swin}, we introduce four different versions of Video Swin Transformer.
The architecture hyper-parameters of these model variants are:
\begin{itemize}
    \item Swin-T: $C=96$, layer numbers = $\{2, 2, 6, 2\}$
    \item Swin-S: $C=96$, layer numbers =$\{2, 2, 18, 2\}$
    \item Swin-B: $C=128$, layer numbers =$\{2, 2, 18, 2\}$
    \item Swin-L: $C=192$, layer numbers =$\{2, 2, 18, 2\}$ 
\end{itemize}
where $C$ denotes the channel number of the hidden layers in the first stage. 
These four versions are about $0.25\times$, $0.5\times$, $1\times$ and $2\times$ the base model size and computational complexity, respectively. The window size is set to $P=8$ and $M=7$ by default. The query dimension of each head is $d=32$, and the expansion layer of each MLP is set to $\alpha=4$.

\subsection{Initialization from Pre-trained Model}

As our architecture is adapted from Swin Transformer~\cite{liu2021swin}, our model can be initialized by its strong pre-trained model on a large-scale dataset.
Compared to the original Swin Transformer, only two building blocks in Video Swin Transformers have different shapes, the linear embedding layer in the first stage and the relative position biases in the Video Swin Transformer block.

For our model, the input token is inflated to a temporal dimension of 2, thus the shape of the linear embedding layer becomes 96$\times$C from 48$\times$C in the original Swin. Here, we directly duplicate the weights in the pre-trained model twice and then multiply the whole matrix by $0.5$ to keep the mean and variance of the output unchanged.
The shape of the relative position bias matrix is ($2P-1$, $2M-1$, $2M-1$), compared to ($2M-1$, $2M-1$) in the original Swin. To make the relative position bias the same within each frame, we duplicate the matrix in the pre-trained model $2P-1$ times to obtain a shape of ($2P-1$, $2M-1$, $2M-1$) for initialization.

\begin{table}[t]
\caption{Comparison to state-of-the-art on Kinetics-400. "384$\uparrow$" signifies that the model uses a larger spatial resolution of 384$\times$384. “Views” indicates \# temporal clip $\times$ \# spatial crop. The magnitudes are Giga ($10^{9}$) and Mega ($10^{6}$) for FLOPs and Param respectively.}
\centering
  \begin{tabular}{l|c|cc|c|cc}
\Xhline{1.0pt}
  Method & Pretrain & Top-1 & Top-5 & Views & FLOPs & Param  \\
  \hline
  R(2+1)D~\cite{tran2018closer} & - & 72.0 & 90.0 & 10 × 1 & 75 & 61.8 \\
  I3D~\cite{carreira2017quo} & ImageNet-1K & 72.1 & 90.3 & - & 108 & 25.0 \\
  NL I3D-101~\cite{wang2018non} & ImageNet-1K & 77.7 & 93.3 & 10 × 3 & 359 & 61.8\\
  ip-CSN-152~\cite{tran2019video} & - & 77.8 & 92.8 & 10 × 3 & 109 & 32.8 \\
  CorrNet-101~\cite{wang2020video} & - & 79.2 & - & 10 × 3 & 224 & - \\
  SlowFast R101+NL~\cite{feichtenhofer2019slowfast} & - & 79.8 & 93.9 & 10 × 3 & 234 & 59.9 \\
  X3D-XXL~\cite{feichtenhofer2020x3d} & - & 80.4 & 94.6 & 10 × 3 & 144 &20.3 \\
  \hline
  MViT-B, 32×3~\cite{fan2021multiscale} & - & 80.2 & 94.4 & 1 × 5 & 170 & 36.6 \\
  MViT-B, 64×3~\cite{fan2021multiscale} & - & 81.2 & 95.1 & 3 × 3 & 455 & 36.6\\
  TimeSformer-L~\cite{timesformer2021} & ImageNet-21K & 80.7 & 94.7 & 1 × 3 & 2380 & 121.4\\
  ViT-B-VTN~\cite{neimark2021VTN} & ImageNet-21K & 78.6 & 93.7 & 1 × 1 &4218 & 11.04\\
  ViViT-L/16x2~\cite{arnab2021vivit} & ImageNet-21K & 80.6 & 94.7 & 4 × 3 &1446 & 310.8\\
  ViViT-L/16x2 320~\cite{arnab2021vivit} & ImageNet-21K & 81.3 & 94.7 & 4 × 3 & 3992 & 310.8\\
  \demph{ip-CSN-152~\cite{tran2019video}} & \demph{IG-65M} & \demph{82.5} & \demph{95.3} & \demph{10 × 3} & \demph{109} & \demph{32.8} \\
  \demph{ViViT-L/16x2~\cite{arnab2021vivit}} & \demph{JFT-300M} & \demph{82.8} & \demph{95.5} & \demph{4 × 3} & \demph{1446} & \demph{310.8} \\
  \demph{ViViT-L/16x2 320~\cite{arnab2021vivit}} & \demph{JFT-300M} & \demph{83.5} & \demph{95.5} & \demph{4 × 3} &\demph{3992} & \demph{310.8}\\
  \demph{ViViT-H/16x2~\cite{arnab2021vivit}} & \demph{JFT-300M} & \demph{84.8} & \demph{95.8} & \demph{4 × 3} & \demph{8316}& \demph{647.5}\\
  \hline
  Swin-T & ImageNet-1K & 78.8 & 93.6 & 4 × 3 & 88 & 28.2 \\
  Swin-S & ImageNet-1K & 80.6 & 94.5 & 4 × 3 & 166 & 49.8 \\
  Swin-B & ImageNet-1K & 80.6 & 94.6 & 4 × 3 & 282 & 88.1 \\
  Swin-B & ImageNet-21K & 82.7 & 95.5 & 4 × 3 & 282 & 88.1 \\ 
  Swin-L & ImageNet-21K & 83.1 & 95.9 & 4 × 3 & 604 & 197.0 \\
  Swin-L (384$\uparrow$) & ImageNet-21K & 84.6 & 96.5 & 4 × 3 & 2107 & 200.0 \\ 
  Swin-L (384$\uparrow$) & ImageNet-21K & \textbf{84.9} & \textbf{96.7} & 10 × 5 & 2107 & 200.0 \\
\Xhline{1.0pt}
  \end{tabular}
\label{tab:k400}
\end{table}

\section{Experiments}

\subsection{Setup}
\label{subsec:setup}
\paragraph{Datasets} For human action recognition, we adopt two versions of the widely-used Kinetics \cite{kay2017kinetics} dataset, Kinetics-400 and Kinetics-600.
Kinetics-400 (K400) consists of $\sim$240k training videos and 20k validation videos in 400 human action categories. 
Kinetics-600 (K600) is an extension of K400 that contains $\sim$370k training videos and 28.3k validation videos from 600 human action categories. 
For temporal modeling, we utilize the popular Something-Something V2 (SSv2)~\cite{goyal2017ssv2} dataset, which consists of 168.9K training videos and 24.7K validation videos over 174 classes.
For all methods, we follow prior art by reporting top-1 and top-5 recognition accuracy.

\paragraph{Implementation Details}
For K400 and K600, we employ an AdamW~\cite{kingma2014adam} optimizer for 30 epochs using a cosine decay learning rate scheduler and 2.5 epochs of linear warm-up. A batch size of 64 is used. As the backbone is initialized from the pre-trained model but the head is randomly initialized, we find that multiplying the backbone learning rate by 0.1 improves performance (shown in Tab.~\ref{tab:lr}). 
Specifically, the initial learning rates for the ImageNet pre-trained backbone and randomly initialized head are set to 3e-5 and 3e-4, respectively.
Unless otherwise mentioned, for all model variants, we sample a clip of 32 frames from each full length video using a temporal stride of 2 and spatial size of $224\times224$, resulting in 16$\times$56$\times$56 input 3D tokens.
Following~\cite{liu2021swin}, an increasing degree of stochastic depth~\cite{huang2016deep} and weight decay is employed for larger models, i.e. $0.1, 0.2, 0.3$ stochastic depth rate and $0.02, 0.02, 0.05$ weight decay for Swin-T, Swin-S, and Swin-B, respectively.
For inference, we follow~\cite{arnab2021vivit} by using $4\times3$ views, where a video is uniformly sampled in the temporal dimension as 4 clips, and for each clip, the shorter spatial side is scaled to 224 pixels and we take 3 crops of size $224\times224$ that cover the longer spatial axis. The final score is computed as the average score over all the views.

For SSv2, we employ an AdamW~\cite{kingma2014adam} optimizer for longer training of 60 epochs with 2.5 epochs of linear warm-up. The batch size, learning rate and weight decay are the same as that for Kinetics. We follow~\cite{mvit2021} by employing a stronger augmentation, including label smoothing, RandAugment~\cite{cubuk2020randaugment}, and random erasing~\cite{zhong2020random}. We also employ stochastic depth~\cite{huang2016deep} with ratio of $0.4$. As also done in~\cite{mvit2021}, we use the model pre-trained on Kinetics-400 as initialization and a window size in temporal dimension of 16 is used. For inference, the final score is computed as the average score of $1\times3$ views.

\begin{table}[t]
\caption{Comparison to state-of-the-art on Kinetics-600.}
\centering
  \begin{tabular}{l|c|cc|c|cc}
\Xhline{1.0pt}
  Method & Pretrain & Top-1 & Top-5 & Views & FLOPs & Param  \\
  \hline
  SlowFast R101+NL~\cite{feichtenhofer2019slowfast} & - & 81.8 & 95.1 & 10 × 3 & 234 & 59.9 \\
  X3D-XL~\cite{feichtenhofer2020x3d} & - & 81.9 & 95.5 & 10 × 3 & 48 & 11.0 \\
  MViT-B-24, 32×3~\cite{mvit2021} & - & 83.8 & 96.3 & 5 × 1 & 236 & 52.9 \\
  TimeSformer-HR~\cite{timesformer2021} & ImageNet-21K & 82.4 & 96 & 1 × 3 & 1703 & 121.4\\
  ViViT-L/16x2 320~\cite{arnab2021vivit} & ImageNet-21K & 83.0 & 95.7 & 4 × 3 & 3992& 310.8\\
  \demph{ViViT-H/16x2~\cite{mvit2021}} & \demph{JFT-300M} & \demph{85.8} & \demph{96.5} & \demph{4 × 3} &\demph{8316} & \demph{647.5} \\
  \hline
  Swin-B & ImageNet-21K & 84.0 & 96.5 & 4 × 3 & 282 & 88.1\\
  Swin-L (384$\uparrow$) & ImageNet-21K & {85.9} & {97.1} & 4 × 3 & 2107 & 200.0 \\
  Swin-L (384$\uparrow$) & ImageNet-21K & \textbf{86.1} & \textbf{97.3} & 10 × 5 & 2107 & 200.0   \\
\Xhline{1.0pt}
  \end{tabular}
\label{tab:k600}
\end{table}

\begin{table}[t]
\caption{Comparison to state-of-the-art on Something-Something v2.}
\centering
  \begin{tabular}{l|c|cc|c|cc}
  \Xhline{1.0pt}
  Method & Pretrain & Top-1 & Top-5 & Views & FLOPs & Param  \\
  \hline
  TimeSformer-HR~\cite{timesformer2021} & ImageNet-21K & 62.5 & - & 1 × 3 & 1703 & 121.4 \\
  SlowFast R101, 8×8~\cite{feichtenhofer2019slowfast} & Kinetics-400 & 63.1 & 87.6 & 1 × 3 & 106 & 53.3 \\
  TSM-RGB~\cite{lin2019tsm} & Kinetics-400 & 63.3 & 88.2 & 2 × 3 & 62 & 42.9\\
  MSNet~\cite{kwon2020motionsqueeze}	& ImageNet-21K & 64.7 & 89.4 & 1 × 1 & 67&24.6 \\
  TEA~\cite{li2020tea} & ImageNet-21K & 65.1 & 89.9 & 10 × 3 & 70& -\\
  blVNet~\cite{fan2019more} & SSv2 & 65.2 & 90.3 & 1 × 1 & 129 & 40.2\\
  ViViT-L/16x2~\cite{arnab2021vivit} & - & 65.4 & 89.8 & - & 903 & 352.1\\
  MViT-B, 64×3~\cite{fan2021multiscale} & Kinetics-400 & 67.7 & 90.9 & 1 × 3 & 455 & 36.6 \\
  \demph{MViT-B-24, 32×3~\cite{fan2021multiscale}} & \demph{Kinetics-600} & \demph{68.7} & \demph{91.5} & \demph{1 × 3} & \demph{236} & \demph{53.2}\\
  \hline
  Swin-B & Kinetics-400 & \textbf{69.6} & \textbf{92.7} & 1 × 3 & 321 & 88.8 \\
  \Xhline{1.0pt}
  \end{tabular}
\label{tab:ssv2}
\end{table}

\subsection{Comparison to state-of-the-art}
\paragraph{Kinetics-400} Table~\ref{tab:k400} presents comparisons to the state-of-the-art backbones, including both convolution-based and Transformer-based on Kinetics-400.
Compared to the state-of-the-art vision Transformers without large-scale pre-training, Swin-S with ImageNet-1K pre-training achieves slightly better performance than MViT-B (32$\times$3)~\cite{mvit2021} which is trained from scratch with similar computation costs. Compared to the state-of-the-art ConvNet X3D-XXL~\cite{feichtenhofer2020x3d}, Swin-S also outperforms it with similar computation costs and fewer views for inference.
For Swin-B, the ImageNet-21K pre-training brings a 2.1\% gain over training on ImageNet-1K from scratch. 
With ImageNet-21K pre-training, our Swin-L (384$\uparrow$) outperforms ViViT-L (320) by 3.3\% on top-1 accuracy with about half less computation costs.
Pre-training on a significantly smaller dataset (ImageNet-21K) than ViViT-H (JFT-300M), our Swin-L (384$\uparrow$) achieves the state-of-the-art performance of 84.9\% on K400.

\paragraph{Kinetics-600} Results on K600 are shown in Table~\ref{tab:k600}. The observations on K600 is similar to those for K400. Compared with the state-of-the-art with ImageNet-21K pre-training, our Swin-L (384$\uparrow$) outperforms ViViT-L (320) by 2.9\% on top-1 accuracy with about half less computation costs.
With pre-training on a significantly smaller dataset (ImageNet-21K) than ViViT-H (JFT-300M), our Swin-L (384$\uparrow$) obtains state-of-the-art accuracy of 86.1\% on K600.

\paragraph{Something-Something v2} Table~\ref{tab:ssv2} compares our approach with the state-of-the-art on SSv2. We follow MViT~\cite{mvit2021} by using the K400 pre-trained model as initialization. With pre-trained models on K400, Swin-B attains 69.6\% top-1 accuracy, surpassing the previous best approach MViT-B-24 with K600 pre-training by 0.9\%. Our approach could be further improved via using larger model (e.g. Swin-L), larger resolution of input (e.g. 384$^2$) and better pre-trained model (e.g. K600). We leave these attempts as future work.

\subsection{Ablation Study}
\paragraph{Different designs for spatiotemporal attention} We ablate three major designs for spatiotemporal attention: joint, split and factorized variants. The joint version jointly computes spatiotemporal attention in each 3D window-based MSA layer, which is our default setting. The split version adds two temporal transformer layers on top of the spatial-only Swin Transformer, which is shown to be effective in ViViT~\cite{arnab2021vivit} and VTN~\cite{neimark2021VTN}. The factorized version adds a temporal-only MSA layer after each spatial-only MSA layer in Swin Transformer, which is found to be effective in TimeSformer~\cite{timesformer2021}.
For the factorized version, to reduce the bad effects of adding randomly initialized layers into the backbone with pre-trained weights, we add a weighting parameter at the end of each temporal-only MSA layer which is initialized as zero.

\begin{table}[h]
\caption{Ablation study on different designs for spatiotemporal attention with Swin-T on K400.}
\centering
  \begin{tabular}{l|cc|cc}
  \Xhline{1.0pt}
   & Top-1 & Top-5 & FLOPs & Param  \\
  \hline
  joint  & 78.8 & 93.6 & 88 & 28.2 \\
  split & 76.4 & 92.1 & 83 & 42.0 \\
  factorized & 78.5 & 93.5 & 95 & 36.5 \\
  \Xhline{1.0pt}
  \end{tabular}
  \label{tab:variants}
\end{table}

Results are shown in Table~\ref{tab:variants}. We can observe that the joint version achieves the best speed-accuracy tradeoff. This is mainly because locality in the spatial domain reduces computation for the joint version while maintaining effectiveness. In contrast, a joint version based on ViT/DeiT would be too computationally expensive.
The split version does not work well in our scenarios. Though this version could naturally benefit from the pre-trained model, the temporal modeling of this version is not as efficient.
The factorized version yields relatively high top-1 accuracy but requires many more parameters than the joint version. This is due the factorized version having a temporal-only attention layer after each spatial-only attention layer, while the joint version performs spatial and temporal attention in the same attention layer.

\begin{table}[h]
\caption{Ablation study on temporal dimension of 3D tokens and temporal window size with Swin-T on K400.}
\centering
  \begin{tabular}{cc|cc|cc}
  \Xhline{1.0pt}
  temporal dimension & Window size & Top 1 & Top 5 & FLOPs & Param  \\
  \hline
  16 & 16$\times$7$\times$7 & 79.1 & 93.8  & 106 & 28.5 \\
  8 & 8$\times$7$\times$7 & 78.5 & 93.2 & 44 & 28.2 \\
  4 & 4$\times$7$\times$7 & 76.7 & 92.5 & 20 & 28.0 \\
  \hline
  16 & 16$\times$7$\times$7 & 79.1 & 93.8 & 106 & 28.5 \\
  16 & 8$\times$7$\times$7 & 78.8 & 93.6 & 88 & 28.2 \\
  16 & 4$\times$7$\times$7 & 78.6 & 93.4 & 79 & 28.0 \\

  \Xhline{1.0pt}
  \end{tabular}
  \label{tab:frame}
\end{table}

\paragraph{Temporal dimension of 3D tokens} We perform an ablation study on the temporal dimension of 3D tokens in a temporally global fashion, where the temporal dimension of 3D tokens is equal to the temporal window size. Results with Swin-T on K400 are shown in Table~\ref{tab:frame}. In general, a larger temporal dimension leads to a higher top-1 accuracy but with greater computation costs and slower inference.

\paragraph{Temporal window size} Fixing the temporal dimension of 3D tokens to 16, we perform an ablation study over temporal window sizes of 4/8/16. Results with Swin-T on K400 are shown in Table~\ref{tab:frame}. We observe that Swin-T with a temporal window size of 8 incurs only a small performance drop of 0.3 compared to a temporal window size of 16 (temporally global), but with a 17\% relative decrease in computation (88 vs.~106). This indicates that temporal locality brings an improved speed-accuracy tradeoff for video recognition. If the number of input frames is extremely large, temporal locality would have an even greater impact.

\begin{table}[h]
    \centering
    \caption{Ablation study on the 3D shifted window approach with Swin-T on K400.}
\centering
  \begin{tabular}{l|cc}
  \Xhline{1.0pt}
  & Top-1 & Top-5 \\
  \hline
  w. 3D shifting & 78.8 & 93.6 \\
  w/o temporal shifting & 78.5 & 93.5 \\
  w/o 3D shifting & 78.1 & 93.3 \\
  \Xhline{1.0pt}
  \end{tabular}
  \label{tab:shift}
  \end{table}

\paragraph{3D shifted windows} Ablations of the \emph{3D shifted windowing} approach on Swin-T are reported for K400 in Table~\ref{tab:shift}. 3D shifted windows bring +0.7\% in top-1 accuracy, and temporally shifted windows yield +0.3\%. The results indicate the effectiveness of the 3D shifted windowing scheme to build connections among non-overlapping windows.

\paragraph{Ratio of backbone/head learning rate} An interesting finding on the ratio of backbone and head learning rates is shown in Table~\ref{tab:lr}.
With a model pre-trained on ImageNet-1K/ImageNet-21K, we observe that a lower learning rate of the backbone architecture (e.g. 0.1$\times$) relative to that of the head, which is randomly initialized, brings gains in top-1 accuracy for K400. 
Also, using the model pre-trained on ImageNet-21K benefits more from this technique, due to the model pre-trained on ImageNet-21K being stronger. 
As a result, the backbone forgets the pre-trained parameters and data slowly while fitting the new video input, leading to better generalization. This observation suggests a direction for further study on how to better utilize pre-trained weights.

\begin{table}[h]
    \centering
    \caption{Ablation study on the ratio of backbone lr and head lr with Swin-B on K400.}
\centering
  \begin{tabular}{cc|cc}
  \Xhline{1.0pt}
  ratio  & Pretrain & Top-1 & Top-5 \\
  \hline
   0.1$\times$ & ImageNet-1K & 80.6 & 94.6\\
   1.0$\times$ & ImageNet-1K & 80.2 & 94.2\\
   0.1$\times$ & ImageNet-21K & 82.6 & 95.7\\
   1.0$\times$ & ImageNet-21K & 82.0 & 95.3\\
  \Xhline{1.0pt}
  \end{tabular}
  \label{tab:lr}
\end{table}

\paragraph{Initialization on linear embedding layer and 3D relative position bias matrix} In ViViT~\cite{arnab2021vivit}, center initialization of the linear embedding layer outperforms inflate initialization by a large margin. This motivates us to conduct an ablation study on these two initialization methods for Video Swin Transformer. 
As shown in Table~\ref{table:init}, we surprisingly find that Swin-T with center initialization obtains the same performance as Swin-T with inflate initialization, of 78.8\% top-1 accuracy using the ImageNet-1K pre-trained model\footnote{As this observation is inconsistent with that in~\cite{arnab2021vivit}, we will analyze the difference once the code of ViViT is released.} On K400.
In this paper, we adopt the conventional inflate initialization on the linear embedding layer by default.

\begin{table}[h]
\begin{minipage}{0.45\linewidth}
\centering
    \caption{Ablation study on the two initialization methods of linear embedding layer with Swin-T on K400.}
  \begin{tabular}{l|cc}
  \Xhline{1.0pt}
  Initialization & Top 1 & Top 5\\
  \hline
  Inflate & 78.8 & 93.6 \\
  Center & 78.8 & 93.7 \\
  \Xhline{1.0pt}
  \end{tabular}
  \label{table:init}
\end{minipage}\hfill
\begin{minipage}{0.45\linewidth}
\centering
    \caption{Ablation study on the two initialization methods of 3D relative position bias matrix with Swin-T on K400.}
  \begin{tabular}{l|cc}
  \Xhline{1.0pt}
  Initialization & Top 1 & Top 5\\
  \hline
  Duplicate & 78.8 & 93.6 \\
  Center & 78.8 & 93.6 \\
  \Xhline{1.0pt}
  \end{tabular}
  \label{table:init-pos}
\end{minipage}
\end{table}

For the 3D relative position bias matrix, we also have two different initialization choices, duplicate or center initialization. Unlike the center initialization method for linear embedding layer, we initialize the 3D relative position bias matrix by masking the relative position bias across different frames with a small negative value (e.g. -4.6), so that each token only focuses inside the same frame from the very beginning. As shown in Table~\ref{table:init-pos}, we find that both initialization methods achieve the same top-1 accuracy of 78.8\% with Swin-T on K400. We adopt duplicate initialization on the 3D Relative Position Bias matrix by default.

\section{Conclusion}
We presented a pure-transformer architecture for video recognition that is based on spatiotemporal locality inductive bias. This model is adapted from the Swin Transformer for image recognition, and thus it could leverage the power of the strong pre-trained image models. The proposed approach achieves state-of-the-art performance on three widely-used benchmarks, Kinetics-400, Kinetics-600 and Something-Something v2. We made the code publicly available to facilitate future study in this field.

\newpage

\bibliographystyle{apalike}
\bibliography{ref}

\end{document}